%% file: main.tex
\documentclass{article} 
\pdfoutput=1
\usepackage{nips15submit_e,times}
\usepackage{url}

\usepackage{tikz, pgf}
\usetikzlibrary{calc}
\usetikzlibrary{decorations.pathreplacing}
\usepackage{amsmath, amsthm, bm, amssymb, dsfont}

\DeclareMathOperator*{\argmax}{arg\,\!max}
\usepackage{algpseudocode, algorithm}
\usepackage{graphicx}
\usepackage{color}
\usepackage{multirow}
\usepackage{wrapfig}
\usepackage{graphicx} 
\definecolor{DarkBlue}{rgb}{0, 0, .4}

\usepackage{caption}
\usepackage{subcaption}
\usepackage[sort&compress,numbers]{natbib}
\usepackage[colorlinks=true,linkcolor=DarkBlue,citecolor=DarkBlue,pdfborder={0 0 0}]{hyperref}

\allowdisplaybreaks

\newcommand{\tr}[1]{ { \ensuremath{\mathrm{tr}\left[ { #1 } \right]} } }

\makeatletter
\newcommand\footnoteref[1]{\protected@xdef\@thefnmark{\ref{#1}}\@footnotemark}
\makeatother

\title{Streaming, Distributed Variational Inference for Bayesian Nonparametrics}
\author{
Trevor Campbell$^1$ \quad Julian Straub$^2$ \quad John W.~Fisher III$^2$ \quad Jonathan
P.~How$^1$\\
$^1$LIDS, $^2$CSAIL, MIT\\
\texttt{\{tdjc@ , jstraub@csail.~, fisher@csail.~, jhow@\}mit.edu} \\
}

\nipsfinalcopy 

\allowdisplaybreaks

\begin{document}

\maketitle

\begin{abstract}
This paper presents a methodology for creating streaming, distributed 
inference algorithms for Bayesian nonparametric (BNP) models.
In the proposed framework, processing nodes receive a sequence of data
minibatches, compute a variational posterior for each, and make 
asynchronous streaming updates to a central model.
In contrast to previous algorithms, the proposed framework is truly 
streaming, distributed, asynchronous, learning-rate-free, 
and truncation-free. 
The key challenge in developing the framework, arising from the fact that BNP
models do not impose an inherent ordering on their components, is finding the correspondence
between minibatch and central BNP posterior components before performing each update.
To address this, the paper develops a combinatorial optimization problem over
component correspondences, and provides an efficient solution
technique.
%
%
%
The paper concludes
with an application of the methodology to
the DP mixture model, with experimental results demonstrating its
practical scalability and performance.
\end{abstract}
\input{intro}
\input{bnpmerge_dp}

\input{reglb}
\input{dpexpt}

\input{conclusion}

\subsubsection*{Acknowledgments}
\vspace{-.1cm}
This work was supported by the Office of Naval Research under ONR MURI
grant N000141110688.

\newpage
\begingroup
\renewcommand{\section}[2]{\subsubsection*{#2}}
{\small
\bibliographystyle{unsrtnat}
\bibliography{main}
}
\endgroup

\end{document}

%% file: intro.tex
\section{Introduction}
Bayesian nonparametric (BNP) stochastic processes are \emph{streaming} priors -- their
unique feature is that they specify, in a probabilistic sense, that the
complexity of a latent model should grow as the amount of observed data increases.
This property captures common sense
in many data analysis problems -- for
example, one would expect to encounter far more topics in a document corpus
after reading $10^6$ documents than after reading $10$ -- and
becomes crucial in settings with unbounded, persistent streams of data. While
their fixed, parametric cousins can be used to infer 
model complexity for datasets with known magnitude a priori~\citep{Nobile94,
Miller13_NIPS}, such priors
are silent with respect to notions of model complexity growth in
streaming data settings.

Bayesian nonparametrics are also naturally
suited to \emph{parallelization} of data processing,
due to the exchangeability, and thus conditional
independence, they often exhibit via de Finetti's theorem. For example,
labels from the Chinese Restaurant process~\citep{Teh10_EML} are rendered
i.i.d.~by conditioning on the underlying Dirichlet process (DP) random measure,
and feature assignments from the Indian Buffet
process~\citep{Griffiths05_NIPS} are rendered
i.i.d.~by conditioning on the underlying beta process (BP) random measure.


Given these properties, one might expect there to be a wealth of inference algorithms
for BNPs that address the challenges associated with
parallelization and streaming. However, previous work 
has only addressed these two settings in concert for parametric
models~\citep{Broderick13_NIPS, Campbell14_UAI},
and only recently has each been addressed individually for BNPs.
In the streaming setting, \citep{Lin13_NIPS} and \citep{Nott14_JCGS} developed
streaming inference for DP mixture models using sequential variational approximation. Stochastic
variational inference~\citep{Hoffman13_JMLR} and related
methods~\citep{Wang11_AISTATS, Bryant09_NIPS, Wang12_NIPS, Hughes13_NIPS}
are often considered streaming algorithms, but their performance depends
on the choice of a learning rate and on the dataset having known, fixed size a priori~\citep{Broderick13_NIPS}. 
Outside of variational approaches, which
are the focus of the present paper, there exist exact parallelized MCMC methods
for BNPs~\citep{Chang13_NIPS,Neiswanger14_UAI}; 
the tradeoff in using such methods is that they provide samples from 
the posterior rather than the distribution itself, and results regarding assessing
convergence remain limited. 
Sequential particle filters for inference have also been
developed~\citep{Carvalho10_BA}, but these suffer issues 
with particle degeneracy and exponential forgetting.

The main challenge posed by the streaming, distributed setting for BNPs is the combinatorial problem of \emph{component identification}. Most BNP models contain some
notion of a countably infinite set of latent ``components'' (e.g.~clusters in a
DP mixture model), and do not impose an inherent ordering on the components.  Thus, in order to 
combine information about the components from multiple processors, the correspondence between 
components must first be found. Brute force search is intractable even for
moderately sized models -- there are $K_1+K_2 \choose K_1$ possible correspondences 
for two sets of components of sizes $K_1$ and $K_2$. Furthermore, there does not yet exist 
a method to evaluate the quality of a component correspondence for BNP models.
This issue has been studied before in the MCMC literature, where it is known as
the ``label switching problem'', but past solution techniques are generally model-specific 
and restricted to use on very simple mixture models~\citep{Stephens00_JRSSB,Jasra05_SS}.

This paper presents a methodology for creating streaming, distributed 
inference algorithms for Bayesian nonparametric models. In the 
proposed framework (shown for a single node A in Figure \ref{fig:framework}), processing nodes receive a sequence of data
minibatches, compute a variational posterior for each, and make asynchronous streaming
updates to a central model using a mapping obtained from a component identification optimization.
The key contributions of this work are as follows. First, we develop a minibatch
posterior decomposition that motivates a learning-rate-free streaming, distributed framework
suitable for Bayesian nonparametrics. Then, we derive the component identification
optimization problem by maximizing the probability of a component matching.
We show that the BNP prior regularizes model complexity in the
optimization; an interesting side effect of this is that regardless
of whether the minibatch variational inference scheme is truncated, the proposed algorithm
is truncation-free. Finally, we provide an efficiently
computable regularization bound for the Dirichlet process prior based on Jensen's inequality\footnote{Regularization bounds for other
popular BNP priors may be found in the supplement.}. The paper concludes  with applications of the methodology to
the DP mixture model, with experimental results demonstrating 
the scalability and performance of the method in practice.

%

%% file: bnpmerge_dp.tex
\section{Streaming, distributed Bayesian nonparametric inference}\label{subsec:genmerge}
\begin{figure}
  \centering
  \begin{subfigure}[b]{.23\textwidth}
    \captionsetup{font=scriptsize}
  \input{frameworkfig1}
  \caption{Retrieve the data/prior}\label{fig:framework1}
  \end{subfigure}
  \begin{subfigure}[b]{.23\textwidth}
    \captionsetup{font=scriptsize}
  \input{frameworkfig2}
  \caption{Perform inference }\label{fig:framework2}
  \end{subfigure}
  \begin{subfigure}[b]{.23\textwidth}
    \captionsetup{font=scriptsize}
  \input{frameworkfig3}
  \caption{Perform component ID}\label{fig:framework3}
  \end{subfigure}
  \begin{subfigure}[b]{.23\textwidth}
    \captionsetup{font=scriptsize}
  \input{frameworkfig4}
  \caption{Update the model}\label{fig:framework4}
  \end{subfigure}
  \vspace{.1cm}
  \captionsetup{font=small}
  \caption{The four main steps of the algorithm that is run asynchronously on each
  processing node. }\label{fig:framework}
\end{figure}
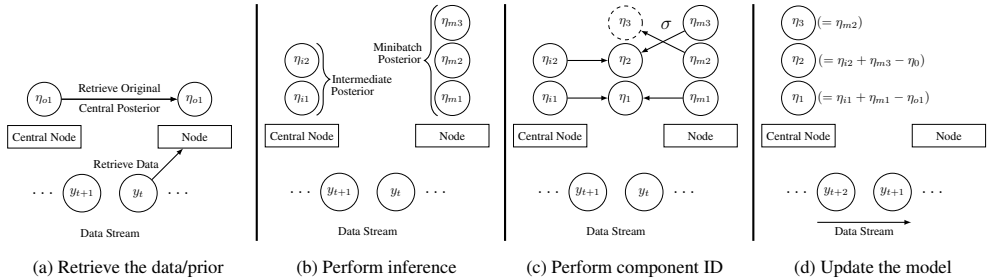
The proposed framework, motivated by a posterior decomposition that will be discussed in
Section \ref{sec:postdecomp}, involves a collection of processing nodes with asynchronous access to a central variational posterior
approximation (shown for a single node in Figure \ref{fig:framework}). Data is provided to each processing node as a sequence
of minibatches. When a processing node receives a minibatch of data, it obtains
the central posterior (Figure \ref{fig:framework1}), and using it as a prior, computes
a minibatch variational posterior approximation (Figure \ref{fig:framework2}).
When minibatch inference is complete, the node then performs 
component identification between the minibatch posterior and 
the current central posterior, accounting for possible modifications made by other
processing nodes (Figure \ref{fig:framework3}). Finally, it merges the minibatch posterior into
the central variational posterior (Figure \ref{fig:framework4}). 

In the following sections, we use the DP
mixture~\citep{Teh10_EML} as a guiding example for the technical development of the inference framework. However,
it is emphasized that the material in this paper generalizes to 
many other BNP models, such as the hierarchical DP (HDP) topic
model~\citep{Teh06_JASA}, BP latent feature model~\citep{DoshiVelez09_ICML}, and
Pitman-Yor (PY)
mixture~\citep{Dubey14_UAI} (see the supplement for further details).


\subsection{Posterior decomposition}\label{sec:postdecomp}
Consider a DP mixture model~\citep{Teh10_EML}, with cluster parameters
$\theta$, assignments $z$, and observed data $y$. 
For each asynchronous update made by each processing node, the dataset is split into three subsets $y = y_o\cup y_i \cup y_m$ for analysis. 
When the processing node receives a \emph{\textbf{m}inibatch} of data
$y_m$, it queries the central processing node for the \emph{\textbf{o}riginal}
posterior $p(\theta, z_o | y_o)$, which will be used as the prior for minibatch
inference. Once inference is complete, it again queries the central processing
node for the \emph{\textbf{i}ntermediate} posterior $p(\theta, z_o, z_i | y_o,
y_i)$ which accounts for asynchronous updates from other processing nodes since
minibatch inference began.
Each subset $y_r$, $r\in\{o, i,
m\}$, has $N_r$ observations $\{y_{rj}\}_{j=1}^{N_r}$, and each variable
$z_{rj}\in\mathbb{N}$ assigns $y_{rj}$ to cluster parameter $\theta_{z_{rj}}$. 
Given the independence of $\theta$ and $z$ in the prior, and the conditional
independence of the data given the latent parameters,
Bayes' rule yields the following  decomposition of the posterior of $\theta$
and $z$ given $y$,
\begin{align}
  \overset{\text{Updated Central
  Posterior}}{\overbrace{\vphantom{p(\theta)^{-1}}p(\theta, z | y)}}\hspace{-.7cm} &\propto 
  \frac{p(z_i, z_m | z_o)}{p(z_i|z_o)p(z_m|z_o)}\cdot \overset{\text{Original
  Posterior}}{\overbrace{\vphantom{p(\theta)^{-1}}p(\theta,
  z_o|y_o)^{-1}}}
  \cdot \overset{\text{Minibatch
  Posterior}}{\overbrace{\vphantom{p(\theta)^{-1}}p(\theta, z_m, z_o|y_m, y_o)}} 
  \cdot \overset{\text{Intermediate Posterior}}{\overbrace{\vphantom{p(\theta)^{-1}}
    p(\theta, z_i, z_o | y_i, y_o)}}.
\label{eq:bnppostcomb}
\end{align}
This decomposition suggests a simple streaming, distributed, asynchronous 
update rule for a processing node: first, obtain the current central posterior
density $p(\theta, z_o | y_o)$, and using it as a prior, compute the minibatch
posterior $p(\theta, z_m, z_o | y_o, y_m)$;
and then update the central posterior density by using (\ref{eq:bnppostcomb})
with the current central posterior density $p(\theta, z_i, z_o | y_i, y_o)$.
However, there are two issues preventing
the direct application of the decomposition rule (\ref{eq:bnppostcomb}):

\textbf{Unknown component correspondence:}
Since it is generally intractable to find the minibatch
posteriors $p(\theta, z_m, z_o | y_o, y_m)$ exactly, approximate methods are required. 
Further, as (\ref{eq:bnppostcomb}) requires the multiplication of
densities, sampling-based methods are difficult to use, suggesting a variational
approach.
Typical mean-field variational techniques
introduce an artificial ordering of the parameters in the
posterior, thereby breaking symmetry that is crucial to combining
posteriors correctly using density multiplication~\citep{Campbell14_UAI}. 
The use of (\ref{eq:bnppostcomb})
with mean-field variational approximations thus requires first solving a component
identification problem.

\textbf{Unknown model size:}
While previous posterior merging procedures required a 1-to-1 matching
between the components of the minibatch posterior and central posterior~\citep{Broderick13_NIPS,Campbell14_UAI}, 
Bayesian nonparametric posteriors break this
assumption. Indeed, the datasets $y_o$, $y_i$, and $y_m$ from the same nonparametric 
mixture model can be generated by the same, disjoint, or an overlapping set of
cluster parameters. In other words, the global number of unique posterior
components cannot be determined until the component identification problem 
is solved and the minibatch posterior is merged.

\subsection{Variational component identification}
Suppose we have the following mean-field exponential family prior and
approximate variational posterior densities in the minibatch decomposition (\ref{eq:bnppostcomb}),
\begin{align}
    p(\theta_k) &= h(\theta_k)e^{\eta_0^TT(\theta_k)-A(\eta_0)} \, \,
    \forall k \in \mathbb{N} \nonumber \\
p(\theta, z_o | y_o) \simeq q_o(\theta, z_o) &= \zeta_o(z_o)\prod_{k=1}^{K_o}
h(\theta_k)e^{\eta_{ok}^TT(\theta_{k}) - A(\eta_{ok})}\nonumber\\
  p(\theta, z_m, z_o | y_m, y_o) \simeq q_m(\theta, z_m, z_o) &=
  \zeta_m(z_m)\zeta_o(z_o)
  \prod_{k=1}^{K_m}
  h(\theta_k)e^{\eta_{mk}^TT(\theta_{k}) - A(\eta_{mk})}\label{eq:varapproxs}\\
  p(\theta, z_i, z_o | y_i, y_o) \simeq q_i(\theta, z_i, z_o) &=
  \zeta_i(z_i)\zeta_o(z_o)
  \prod_{k=1}^{K_i}
  h(\theta_k)e^{\eta_{ik}^TT(\theta_{k}) - A(\eta_{ik})}, \nonumber
\end{align}
where $\zeta_r(\cdot)$, $r\in\{o, i, m\}$ are products of categorical distributions 
for the cluster labels $z_r$, and the goal is to use the posterior decomposition
(\ref{eq:bnppostcomb}) to find the updated posterior approximation
\begin{align}
p(\theta, z | y) \simeq q(\theta, z) &=
  \zeta(z)\prod_{k=1}^Kh(\theta_k)e^{\eta_k^TT(\theta_k) -
  A(\eta_k)}.\label{eq:varapprox2}
\end{align}
As mentioned in the previous section, the artificial ordering of components
causes the na\"{i}ve application of
(\ref{eq:bnppostcomb}) with variational approximations to fail, as
disparate components from the approximate posteriors may be merged erroneously.
This is demonstrated in Figure \ref{fig:synthbroken}, which shows results
from a synthetic experiment (described in Section \ref{sec:expts})
ignoring component identification. As the number of parallel 
threads increases, more matching mistakes are made, leading to decreasing 
model quality.

To address this, first note that there is no issue with the first $K_o$ components of $q_m$
and $q_i$; these can be merged directly since they each correspond 
to the $K_o$ components of $q_o$. 
Thus, the component identification problem reduces to finding the correspondence
between the last $K'_m =
K_m-K_o$ components of the minibatch posterior and the last $K'_i = K_i - K_o$ 
components of the intermediate posterior. For notational simplicity (and without
loss of generality), fix the component ordering of the intermediate posterior $q_i$, and define
$\sigma\! :\! \left[K_m\right] \to \left[K_i+K'_m\right]$ 
to be 
the 1-to-1 mapping from minibatch posterior component $k$ to updated
central posterior component $\sigma(k)$, where $\left[K\right] := \{1, \dots,
K\}$. 
The fact that the first $K_o$ 
components have no ordering ambiguity can be expressed as $\sigma(k) = k
\, \, \forall k\in\left[K_o\right]$. Note that the maximum number of components after
merging is $K_i+K'_m$, since each of the last $K'_m$ components in the minibatch
posterior may correspond to new components in the intermediate posterior. 
After substituting the three variational approximations (\ref{eq:varapproxs}) into
(\ref{eq:bnppostcomb}), the goal of the component identification optimization is
to find the 1-to-1 mapping $\sigma^\star$ that
yields the largest updated posterior normalizing
constant, i.e.~matches components with similar densities,
\begin{align}
  \begin{aligned}
  \sigma^\star \gets \argmax_{\sigma} \quad &\sum_z \! \int_\theta
  \frac{p(z_i, z_m | z_o)}{p(z_i | z_o)p(z_m | z_o)} q_o(\theta, z_o)^{-1}
   q^\sigma_m(\theta,
  z_m, z_o)q_i(\theta, z_i, z_o)\\
  \mathrm{s.t.}\quad & q^\sigma_m(\theta, z_m) = \zeta^\sigma_m(z_m)
  \prod_{k=1}^{K_m}
  h(\theta_{\sigma(k)})  e^{\eta_{mk}^TT(\theta_{\sigma(k)}) - A(\eta_{mk})}\\
  & \sigma(k) = k, \, \, \forall k \in \left[K_o\right], \sigma\text{ 1-to-1}
  \end{aligned}\label{eq:bnpoptpre}
\end{align}
where $\zeta^\sigma_m(z_m)$ is the distribution such that
$\mathrm{P}_{\!\zeta^\sigma_{m}}(z_{mj} = \sigma(k)) =
\mathrm{P}_{\!\zeta_m}(z_{mj} = k)$.  
Taking the logarithm of the objective and exploiting the mean-field decoupling
allows the separation of the objective into a sum of two terms: one expressing
the quality of the matching between components (the integral over $\theta$), and one
that regularizes the final model size (the sum over $z$). While the first term is
available in closed form, the second is in general not.
Therefore, 
using the concavity of the logarithm function, Jensen's inequality yields a
lower bound that can be used in place of the intractable
original objective, resulting in the final component identification optimization:
\begin{align}
  \begin{aligned}
    \sigma^\star \gets \argmax_{\sigma}\quad &
    \sum_{k=1}^{K_i+K'_m} A\left( \tilde{\eta}^\sigma_k \right) +
    \mathbb{E}_\zeta^\sigma\left[\log p(z_i, z_m ,  z_o)\right]\\
    \mathrm{s.t.}\quad & \tilde{\eta}^\sigma_k = \tilde{\eta}_{ik} +
    \tilde{\eta}^\sigma_{mk}
    - \tilde{\eta}_{ok}\\
    &\sigma(k) = k \, \, \forall k \in \left[K_o\right], \sigma\text{ 1-to-1}.
\end{aligned}\label{eq:bnpopt}
\end{align}
A more detailed derivation of the optimization may be found in the supplement.
$\mathbb{E}^\sigma_{\zeta}$ denotes expectation under the distribution
$\zeta_o(z_o)\zeta_i(z_i)\zeta^\sigma_m(z_m)$, and
\begin{align}
\tilde{\eta}_{rk} = \left\{\begin{array}{ll}
                          \eta_{rk} & k \leq K_r\\
                          \eta_0 & k > K_r
                        \end{array}\right.\forall r \in\{o, i, m\}, \quad  
      \tilde{\eta}^\sigma_{mk} = \left\{\begin{array}{ll}
                          \eta_{m\sigma^{\!-1}(k)} & k \in
                          \sigma\left(\left[K_m\right]\right)\\
                          \eta_0 & k \notin\sigma\left(\left[K_m\right]\right)
                        \end{array}\right.,
                        \label{eq:etatilde}
\end{align}
where $\sigma\left(\left[K_m\right]\right)$ denotes the range of the mapping
$\sigma$. The definitions in (\ref{eq:etatilde}) ensure that the prior
$\eta_0$ is used whenever a posterior $r\in\{i, m, o\}$ does not contain
a particular component $k$.  The intuition for the optimization
(\ref{eq:bnpopt}) is that it combines finding component correspondences
with high similarity (via the log-partition function) with
a regularization term\footnote{This is equivalent to the KL-divergence regularization
  $-\mathrm{KL}\left[\zeta_o(z_o)\zeta_i(z_i)\zeta^\sigma_m(z_m)
    \left|\vphantom{\frac{p}{p}}\right| p(z_i, z_m , z_o)\right]$.} on the final updated posterior model size.

Despite its motivation from the Dirichlet process mixture, the component identification optimization (\ref{eq:bnpopt})
is not specific to this model. Indeed, the derivation did not rely on any
properties specific to the Dirichlet process mixture; the optimization applies to any Bayesian nonparametric model with a
set of ``components'' $\theta$, and a set of combinatorial ``indicators'' $z$. For example, the
optimization applies to the hierarchical Dirichlet process topic
model~\citep{Wang11_AISTATS} with topic word distributions $\theta$ and 
local-to-global topic correspondences $z$, and to the beta process latent
feature model~\citep{Griffiths05_NIPS} with features $\theta$ and 
binary assignment vectors $z$. The form of the objective in the component identification optimization (\ref{eq:bnpopt}) reflects this generality. 
In order to apply the proposed streaming, distributed method to a particular model, one simply needs a black-box 
variational inference algorithm that computes posteriors of the form
(\ref{eq:varapproxs}), and a way to compute or bound the expectation in the objective of
(\ref{eq:bnpopt}). 

%
%
%
  
\subsection{Updating the central posterior}
To update the central posterior, the node first locks it and
solves for $\sigma^\star$ via (\ref{eq:bnpopt}). Locking prevents other nodes
from solving (\ref{eq:bnpopt}) or modifying the central posterior, but does not
prevent other nodes from reading the central posterior, obtaining
minibatches, or performing inference; the synthetic experiment in Section
\ref{sec:expts} shows that this does not incur a significant time penalty in practice. Then the processing node transmits $\sigma^\star$ and its minibatch variational
posterior to the central processing node
where the product decomposition (\ref{eq:bnppostcomb}) is used to find the updated
central variational posterior $q$ in (\ref{eq:varapprox2}), with parameters
\begin{align}
  K &= \max \left\{K_i, \max_{k\in\left[K_m\right]}
\sigma^\star(k)\right\}, \quad
  \zeta(z) = \zeta_i(z_i)\zeta_o(z_o)\zeta^{\sigma^\star}_m(z_m), \quad \eta_k = \tilde{\eta}_{ik} + \tilde{\eta}^{\sigma^\star}_{mk} -
  \tilde{\eta}_{ok}. \label{eq:centralupdate}
\end{align}
Finally, the node unlocks the central posterior, and  
the next processing node to receive a new minibatch will use the above $K$, $\zeta(z)$, and $\eta_k$ 
from the central node as their
$K_o$, $\zeta_o(z_o)$, and $\eta_{ok}$. 




%% file: frameworkfig1.tex
\raisebox{.01cm}{\scalebox{.5}{\begin{tikzpicture}
  \tikzstyle{dnode}=[draw, circle, minimum size=1cm];
  \tikzstyle{mnode}=[draw, circle, minimum size=.9cm];
  \node[dnode] (yt) at (0, 0) {$y_t$};
  \node[dnode] (ytn1) at (-1.5, 0) {$y_{t+1}$};
\node[circle, minimum size=1cm] (dotsr) at (1.1, 0) {\Large $\dots$};
  \node[circle, minimum size=1cm] (dotsl) at (-2.5, 0) {\Large$\dots$};
    \node (ds) at ($(yt) + (-.75, -1.05)$) {\small Data Stream};

 \node[draw, rectangle, minimum width=1.95cm, minimum height=.6cm] (proc) at
 ($(yt) + (1.5, 1.5)$) {\small Node}; 

\node[draw, rectangle, minimum width=1.95cm, minimum height=.6cm](srv) at
($(proc) + (-4, 0)$) {\small Central Node};

\node[mnode] (ct1) at ($(srv) + (0, 1)$) {$\eta_{o1}$};

\node[mnode] (pt1) at ($(proc) + (0, 1)$) {$\eta_{o1}$};

\draw[-latex, thick] (ct1) -- (pt1) node[midway, above] {\small Retrieve
Original};
\draw[-latex, thick] (ct1) -- (pt1) node[midway, below] {\small Central Posterior};

\draw[-latex, thick] (yt) -- (proc) node[midway, left, xshift=-.1cm] {\small Retrieve Data};
  \end{tikzpicture}}}

%% file: frameworkfig2.tex
\scalebox{.5}{\begin{tikzpicture}
\tikzstyle{dnode}=[draw, circle, minimum size=1cm];
  \tikzstyle{mnode}=[draw, circle, minimum size=.9cm];
  \node[dnode] (yt) at (0, 0) {$y_t$};
  \node[dnode] (ytn1) at (-1.5, 0) {$y_{t+1}$};
\node[circle, minimum size=1cm] (dotsr) at (1.1, 0) {\Large $\dots$};
  \node[circle, minimum size=1cm] (dotsl) at (-2.5, 0) {\Large$\dots$};

    \node (ds) at ($(yt) + (-.75, -1.05)$) {\small Data Stream};

 \node[draw, rectangle, minimum width=1.95cm, minimum height=.6cm] (proc) at
 ($(yt) + (1.5, 1.5)$) {\small Node }; 

\node[draw, rectangle, minimum width=1.95cm, minimum height=.6cm](srv) at
($(proc) + (-4, 0)$) {\small Central Node};

\node[mnode] (ct1) at ($(srv) + (0, 1)$) {$\eta_{i1}$};
\node[mnode] (ct2) at ($(srv) + (0, 2)$) {$\eta_{i2}$};

\node[mnode] (pt1) at ($(proc) + (0, 1)$) {$\eta_{m1}$};
\node[mnode] (pt2) at ($(proc) + (0, 2)$) {$\eta_{m2}$};
\node[mnode] (pt3) at ($(proc) + (0, 3)$) {$\eta_{m3}$};

\draw[decorate, thick, decoration={brace, amplitude=10pt, raise=4pt}] ( $(pt1.south) + (-.25, 0)$ ) -- ($ (pt3.north) + (-.25, 0) $);
\node (lbl) at ($(pt2) + (-1.45, .35)$) {\small Minibatch};
\node (lbl2) at ($(pt2) + (-1.4, .05)$) {\small Posterior};

\draw[decorate, thick, decoration={brace, mirror, amplitude=10pt,
raise=4pt}] ( $(ct1.south) + (.25, 0)$ ) -- ($ (ct2.north) + (.25, 0) $);
\node (lbl3) at ($(ct2) + (1.6, -.5)$) {\small Intermediate};
\node (lbl4) at ($(ct2) + (1.35, -.8)$) {\small Posterior};

\draw[-, ultra thick] ($(srv) + (-1.2, -2.8)$) -- ($(srv) + (-1.2, 3.5)$);
  \end{tikzpicture}}

%% file: frameworkfig3.tex
\scalebox{.5}{\begin{tikzpicture}
\tikzstyle{dnode}=[draw, circle, minimum size=1cm];
  \tikzstyle{mnode}=[draw, circle, minimum size=.9cm];

  \node[dnode] (yt) at (0, 0) {$y_t$};
  \node[dnode] (ytn1) at (-1.5, 0) {$y_{t+1}$};
\node[circle, minimum size=1cm] (dotsr) at (1.1, 0) {\Large $\dots$};
  \node[circle, minimum size=1cm] (dotsl) at (-2.5, 0) {\Large$\dots$};

    \node (ds) at ($(yt) + (-.75, -1.05)$) {\small Data Stream};

 \node[draw, rectangle, minimum width=1.95cm, minimum height=.6cm] (proc) at
 ($(yt) + (1.5, 1.5)$) {\small Node}; 

\node[draw, rectangle, minimum width=1.95cm, minimum height=.6cm](srv) at
($(proc) + (-4, 0)$) {\small Central Node};

\node[mnode] (ct1) at ($(srv) + (0, 1)$) {$\eta_{i1}$};
\node[mnode] (ct2) at ($(srv) + (0, 2)$) {$\eta_{i2}$};

\node[mnode] (pt1) at ($(proc) + (0, 1)$) {$\eta_{m1}$};
\node[mnode] (pt2) at ($(proc) + (0, 2)$) {$\eta_{m2}$};
\node[mnode] (pt3) at ($(proc) + (0, 3)$) {$\eta_{m3}$};

\node[mnode] (mt1) at ($0.5*(proc)+0.5*(srv) + (0, 1)$) {$\eta_{1}$};
\node[mnode] (mt2) at ($0.5*(proc)+0.5*(srv) + (0, 2)$) {$\eta_{2}$};
\node[mnode, dashed] (mt3) at ($0.5*(proc)+0.5*(srv) + (0, 3)$) {$\eta_{3}$};

%
%
%

\draw[-latex, thick] (ct1) -- (mt1);
\draw[-latex, thick] (pt1) -- (mt1);

\draw[-latex, thick] (ct2) -- (mt2);
\draw[-latex, thick] (pt3) -- (mt2);

\draw[-latex, thick] (pt2) -- (mt3) node[midway, above, yshift=.2cm, xshift=.1cm]{\Large
$\sigma$};


\draw[-, ultra thick] ($(srv) + (-1.2, -2.8)$) -- ($(srv) + (-1.2, 3.5)$);
  \end{tikzpicture}}

%% file: frameworkfig4.tex
\scalebox{.5}{\begin{tikzpicture}
\tikzstyle{dnode}=[draw, circle, minimum size=1cm];
  \tikzstyle{mnode}=[draw, circle, minimum size=.9cm];

  \node[dnode] (yt) at (0, 0) {$y_{t+1}$};
  \node[dnode] (ytn1) at (-1.5, 0) {$y_{t+2}$};
\node[circle, minimum size=1cm] (dotsr) at (1.1, 0) {\Large $\dots$};
  \node[circle, minimum size=1cm] (dotsl) at (-2.5, 0) {\Large$\dots$};

  \draw[-latex, thick] ($(ytn1.south) + (-.5, -.3)$) -- ($(yt.south) + (0.5,
  -.3)$) node[midway, below] {\small Data Stream};

 \node[draw, rectangle, minimum width=1.95cm, minimum height=.6cm] (proc) at
 ($(yt) + (1.5, 1.5)$) {\small  Node}; 

\node[draw, rectangle, minimum width=1.95cm, minimum height=.6cm](srv) at
($(proc) + (-4, 0)$) {\small Central Node};

\node[mnode] (ct1) at ($(srv) + (0, 1)$) {$\eta_1$}; 
\node (eq1) at ($(ct1) + (2, 0)$) {$\left(=\eta_{i1}+\eta_{m1}-\eta_{o1}\right)$};

\node[mnode] (ct2) at ($(srv) + (0, 2)$) {$\eta_2$};
\node (eq2) at ($(ct2) + (1.95, 0)$)
{$\left(=\eta_{i2}+\eta_{m3}-\eta_0\right)$};

\node[mnode] (ct3) at ($(srv) + (0, 3)$) {$\eta_3$};
\node (eq3) at ($(ct3) + (1.1, 0)$) {$\left(=\eta_{m2}\right)$};

\draw[-, ultra thick] ($(srv) + (-1.2, -2.8)$) -- ($(srv) + (-1.2, 3.5)$);

  \end{tikzpicture}}

%% file: reglb.tex
\section{Application to the Dirichlet process mixture model}\label{sec:reglb}
The expectation in the objective of (\ref{eq:bnpopt}) is typically intractable to compute in
closed-form; therefore, a suitable lower bound may be used in its place. 
This section presents such a bound for the Dirichlet process, and discusses the
application of the proposed inference framework to the Dirichlet process mixture
model using the developed bound. Crucially, the lower bound decomposes such that the optimization
(\ref{eq:bnpopt}) becomes a maximum-weight bipartite matching problem. Such
problems are solvable
in polynomial time~\citep{Edmonds72_ACM} by the Hungarian algorithm, leading to a tractable component
  identification step in the proposed streaming, distributed framework.


%
%
%
%

%
%

\subsection{Regularization lower bound}
For the Dirichlet process with concentration parameter $\alpha>0$,
$p(z_i, z_m, z_o)$ is the Exchangeable Partition Probability Function (EPPF)~\citep{Pitman95_PTRF}
\begin{align}
   p(z_i, z_m, z_o)&\propto \alpha^{|\mathcal{K}|-1}\prod_{k\in\mathcal{K}}
   \left(n_k-1\right)!,
 \end{align}
 where $n_k$ is the amount of data assigned to cluster $k$, and
 $\mathcal{K}$ is the set of labels of nonempty clusters. 
Given that the variational distribution $\zeta_r(z_r)$, $r \in \{i, m, o\}$ is a product of 
independent categorical
distributions $\zeta_r(z_r) = \prod_{j=1}^{N_r}
\prod_{k=1}^{K_r}\pi_{rjk}^{\mathds{1}\left[z_{rj}=k\right]}$, 
Jensen's inequality may be used to bound the regularization in (\ref{eq:bnpopt})
below (see the supplement for further details) by 
\begin{align}
  \begin{aligned}
  \mathbb{E}^\sigma_{\zeta}\left[\log p(z_i, z_m, z_o)\right]
  &\geq \sum_{k=1}^{K_i+K'_m}\left(1-e^{\tilde{s}^\sigma_k}\right)\log\alpha +
  \log\Gamma\left(\max\left\{2, \tilde{t}^\sigma_k\right\}\right)
  + C\\
  \tilde{s}^\sigma_k &= \tilde{s}_{ik} +
  \tilde{s}^\sigma_{mk} + \tilde{s}_{ok}, \quad \tilde{t}^\sigma_k = \tilde{t}_{ik} +
  \tilde{t}^\sigma_{mk} + \tilde{t}_{ok},\label{eq:dpreg}
\end{aligned}
\end{align}
where $C$ is a constant with respect to the component mapping $\sigma$, and 
\begin{align}
  \begin{aligned}
    \tilde{s}_{rk} &= \left\{\begin{array}{ll}
                         \scriptstyle \sum_{j=1}^{N_r}\log(1-\pi_{rjk})
                         &\scriptstyle  k \leq K_r\\
                        \scriptstyle 0 &\scriptstyle  k > K_r
                      \end{array}\right.\, \scriptstyle \forall r\in \{o, i, m\} &&
\tilde{t}_{rk} &= \left\{\begin{array}{ll}
                         \scriptstyle \sum_{j=1}^{N_r}\pi_{rjk}
                         &\scriptstyle  k \leq K_r\\
                        \scriptstyle 0 &\scriptstyle  k > K_r
                      \end{array}\right. \scriptstyle \, \forall r\in \{o, i, m\} \\
\tilde{s}^\sigma_{mk} &= \left\{\begin{array}{ll}
      \scriptstyle \sum_{j=1}^{N_m}\log(1-\pi_{mj\sigma^{-1}(k)}) &\scriptstyle  k \in
      \sigma\left(\left[K_m\right]\right)\\
                       \scriptstyle  0 &\scriptstyle  k \notin
      \sigma\left(\left[K_m\right]\right)
    \end{array}\right. &&
    \tilde{t}^\sigma_{mk} &= \left\{\begin{array}{ll}
      \scriptstyle \sum_{j=1}^{N_m}\pi_{mj\sigma^{-1}(k)} & \scriptstyle k \in
      \sigma\left(\left[K_m\right]\right)\\
                       \scriptstyle  0 &\scriptstyle  k \notin
      \sigma\left(\left[K_m\right]\right)
                      \end{array}\right. .
\end{aligned}\label{eq:enkkplus}
\end{align}
 Note that the bound (\ref{eq:dpreg})
allows incremental updates: after finding the optimal mapping $\sigma^\star$, 
the central update (\ref{eq:centralupdate}) can be augmented
by updating the values of $s_k$ and $t_k$ on the central node to
\begin{align}
  s_{k} \gets \tilde{s}_{ik} + \tilde{s}^{\sigma^\star}_{mk} + \tilde{s}_{ok}, \quad
  t_{k} \gets \tilde{t}_{ik} +
  \tilde{t}^{\sigma^\star}_{mk} + \tilde{t}_{ok}. \label{eq:stupdate}
\end{align}
As with $K$, $\eta_k$, and $\zeta$ from (\ref{eq:centralupdate}),
after performing the regularization statistics update (\ref{eq:stupdate}), 
a processing node that receives a new minibatch will use the above $s_k$ and
$t_k$ as their $s_{ok}$ and $t_{ok}$, respectively. 
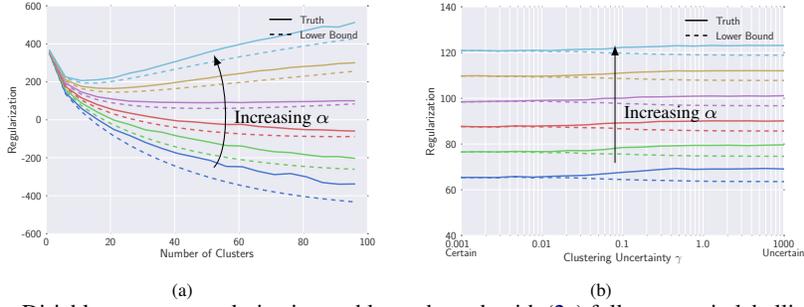
\begin{figure}[t]
  \centering
  \hspace{-.6cm}
  \begin{subfigure}[b]{.35\textwidth}
  \centering
  \captionsetup{font=scriptsize}
  \input{reglbfig1}
  \vspace{-.4cm}
  \caption{}\label{fig:reglbvsk}
  \end{subfigure}
  \hspace{.5cm}
  \begin{subfigure}[b]{.35\textwidth}
  \centering
  \captionsetup{font=scriptsize}
  \input{reglbfig2}
  \vspace{-.4cm}
  \caption{}\label{fig:reglbvsu}
  \end{subfigure}
  \captionsetup{font=small}
  \caption{The Dirichlet process regularization and lower bound, with
    (\ref{fig:reglbvsk}) fully uncertain labelling and varying number of
    clusters, and (\ref{fig:reglbvsu}) the number of clusters fixed with varying
  labelling uncertainty.}\label{fig:reglb}
\end{figure}

Figure \ref{fig:reglb} demonstrates the behavior of the lower bound in a
synthetic experiment with $N=100$ datapoints for various DP concentration
parameter values $\alpha \in \left[10^{-3}, 10^3\right]$. The true
regularization $\log \mathbb{E}_\zeta\left[p(z)\right]$ was computed by sample
approximation with $10^4$
samples. In Figure \ref{fig:reglbvsk}, the
number of clusters $K$ was varied, with symmetric categorical label weights set to
$\frac{1}{K}$. This figure demonstrates two important phenomena. First,
the bound increases as $K\to 0$; in other words, it gives preference to fewer, larger
clusters, which is the typical BNP ``rich get richer'' property. Second, the
behavior of the bound as $K\to N$ depends on the concentration parameter
$\alpha$ -- as $\alpha$ increases, more clusters are preferred. In Figure \ref{fig:reglbvsu}, 
the number of clusters $K$ was fixed to 10, and the categorical label weights were sampled from a symmetric Dirichlet distribution
with parameter $\gamma \in \left[10^{-3}, 10^3\right]$. This figure demonstrates
that the bound does not degrade significantly with high labelling uncertainty,
and is nearly exact for low labelling uncertainty.
Overall, Figure \ref{fig:reglbvsk}
demonstrates that the proposed lower bound exhibits similar behaviors to the true regularization, 
supporting its use in the optimization (\ref{eq:bnpopt}).

\subsection{Solving the component identification
optimization}\label{sec:matchingopt}
Given that both the regularization (\ref{eq:dpreg}) and component matching
score in the objective (\ref{eq:bnpopt}) decompose as a sum of terms for
each $k\in\left[K_i+K'_m\right]$, the objective
can be rewritten using a matrix of matching scores $\mathbf{R}\in
\mathbb{R}^{\left(K_i+K'_m\right)\times \left(K_i+K'_m\right)}$
and selector variables $\mathbf{X}\in \{0, 1\}^{\left(K_i+K'_m\right)\times
\left(K_i+K'_m\right)}$.
Setting $\mathbf{X}_{kj} = 1$ indicates that component $k$ in the minibatch
posterior is matched to component $j$ in the intermediate posterior
(i.e.~$\sigma(k) = j$), providing a score $\mathbf{R}_{kj}$ defined using (\ref{eq:etatilde}) and (\ref{eq:enkkplus}) as
\begin{align}
  \hspace{-.2cm}\mathbf{R}_{kj} &= A\left(\tilde{\eta}_{ij} \!+ \tilde{\eta}_{mk} -
  \tilde{\eta}_{oj}\right)\! + \! \left(1-e^{\tilde{s}_{ij}+\tilde{s}_{mk} +
  \tilde{s}_{oj}}\right)\log\alpha \!+\!
  \log\Gamma\left(\max\left\{2, \tilde{t}_{ij}+\tilde{t}_{mk} +
  \tilde{t}_{oj} \right\}\right).
\end{align}
The optimization
(\ref{eq:bnpopt})  can be rewritten in terms of $\mathbf{X}$ and $\mathbf{R}$ as
\begin{align}
  \begin{aligned}
    \mathbf{X}^\star \gets \argmax_{\mathbf{X}}\quad
    &\tr{\mathbf{X}^T\mathbf{R}}\\
    \mathrm{s.t.}\quad & \mathbf{X}\mathbf{1} = \mathbf{1}, \quad
    \mathbf{X}^T\mathbf{1} = \mathbf{1}, \quad
    \mathbf{X}_{kk} = 1, \forall k \in \left[K_o\right]\\
    &\mathbf{X} \in \{0, 1\}^{\left(K_i+K'_m\right) \times
    \left(K_i+K'_m\right)}, \quad \mathbf{1} = \left[1, \dots, 1\right]^T.
  \end{aligned}
\end{align}
The first two constraints express the 1-to-1 property of $\sigma(\cdot)$. The
constraint $\mathbf{X}_{kk} = 1 \forall k \in \left[K_o\right]$ 
fixes the upper $K_o\times K_o$ block of $\mathbf{X}$ to $\mathbf{I}$ (due to the fact that the 
first $K_o$ components are matched directly), and the
off-diagonal blocks to $\mathbf{0}$. Denoting $\mathbf{X}'$, $\mathbf{R}'$ to be the lower
right $\left(K'_i+K'_m\right) \times
\left(K'_i+K'_m\right)$ blocks of $\mathbf{X}$, $\mathbf{R}$, 
the remaining optimization problem is a linear assignment problem on
$\mathbf{X}'$  with cost matrix $-\mathbf{R}'$, which can be solved
using the Hungarian algorithm\footnote{For the
  experiments in this work, we used the implementation
 at
 \url{github.com/hrldcpr/hungarian}.}. Note that if $K_m = K_o$ or $K_i = K_o$, 
 this implies that no matching problem needs to be solved -- the first $K_o$
 components of the minibatch posterior are matched directly, and the last $K'_m$
 are set as new components. In practical implementation of the framework,
 new clusters are typically discovered at a diminishing rate as more data are observed, 
 so the number of matching problems that are solved likewise tapers off. 
 The final optimal component mapping $\sigma^\star$ is found by 
 finding the nonzero elements of $\mathbf{X}^\star$:
  \begin{align}
    \sigma^\star(k) \gets \argmax_j \mathbf{X}^\star_{kj} \, \, \forall k \in
    \left[K_m\right].
  \end{align}

%% file: reglbfig1.tex
\scalebox{1}{\begin{tikzpicture}
  \node (fig) at (0, 0) {\includegraphics[width=\columnwidth]{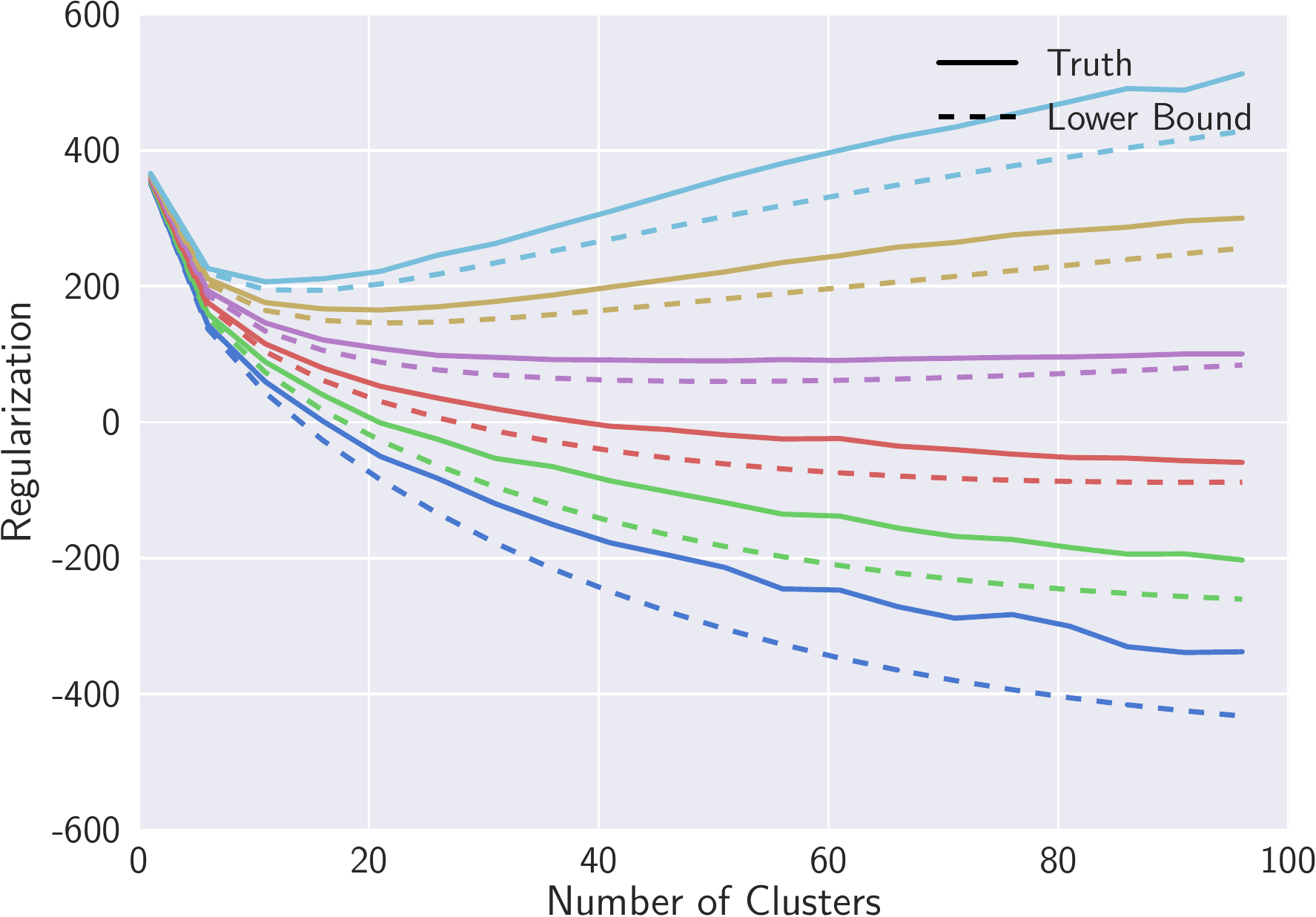}};
  \draw[-latex] (.3, -.5) .. controls (.5, -.3) and (.5, .5) .. (0.3, 1)
  node[right, pos=0.51]{\scriptsize Increasing $ \alpha$};
\end{tikzpicture}}

%% file: reglbfig2.tex
\raisebox{-.08cm}{\scalebox{.98}{\begin{tikzpicture}
  \node (fig) at (0, 0) {\includegraphics[width=1.05\columnwidth]{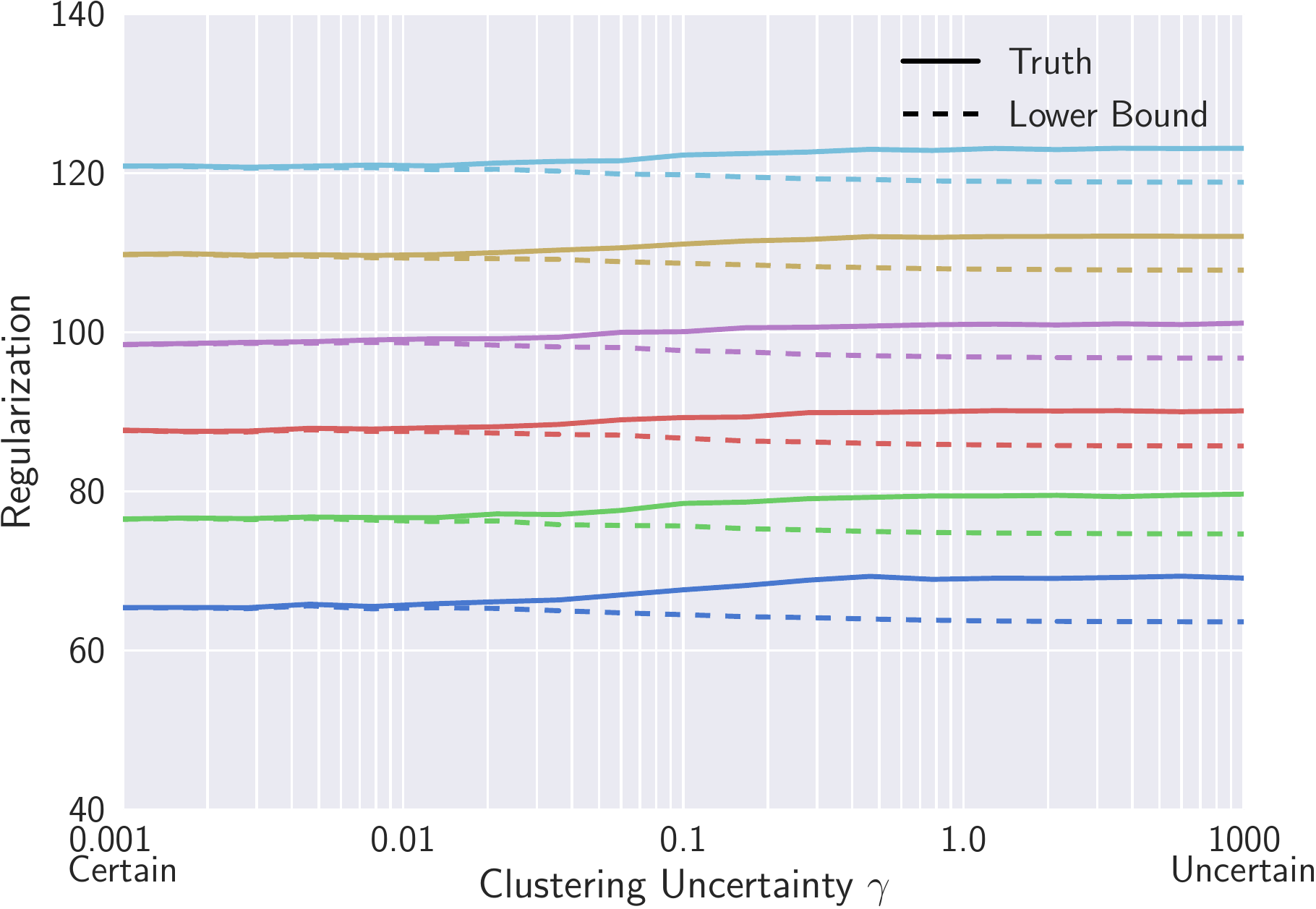}};
  \draw[-latex] (0, -.4) -- (0, 1.2) node[right, pos=0.42]{\scriptsize Increasing $ \alpha$};
\end{tikzpicture}}}

%% file: dpexpt.tex
\section{Experiments}\label{sec:expts}
\begin{figure*}[t!]
\centering
  \begin{subfigure}[b]{.305\textwidth}
  \includegraphics[width=1.1\textwidth]{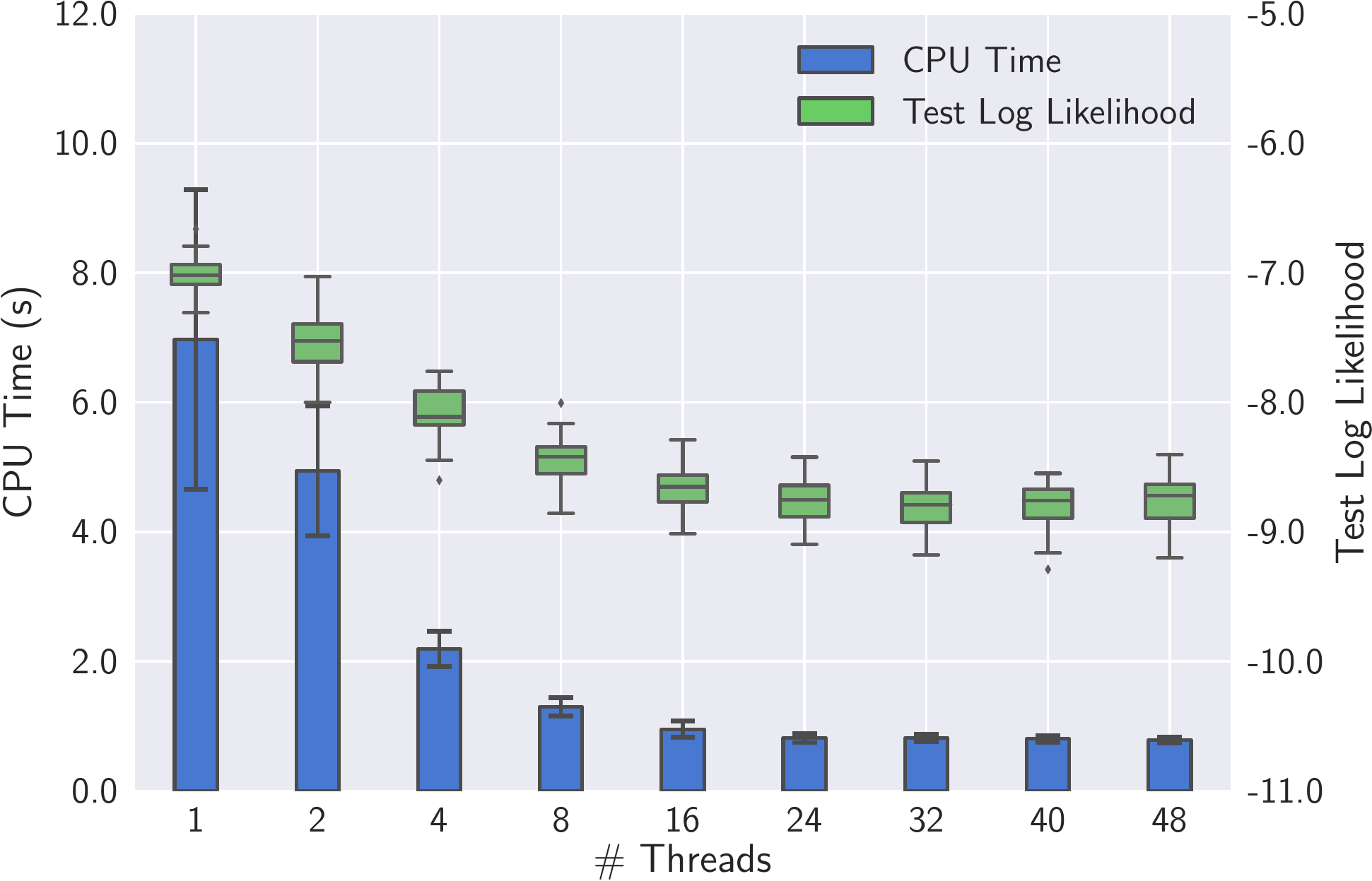}
  \captionsetup{font=scriptsize}
  \caption{SDA-DP without component ID}\label{fig:synthbroken}
  \vspace{.4cm}
\end{subfigure}
\hspace{.3cm}
\begin{subfigure}[b]{0.308\textwidth}
    \includegraphics[width=1.1\textwidth]{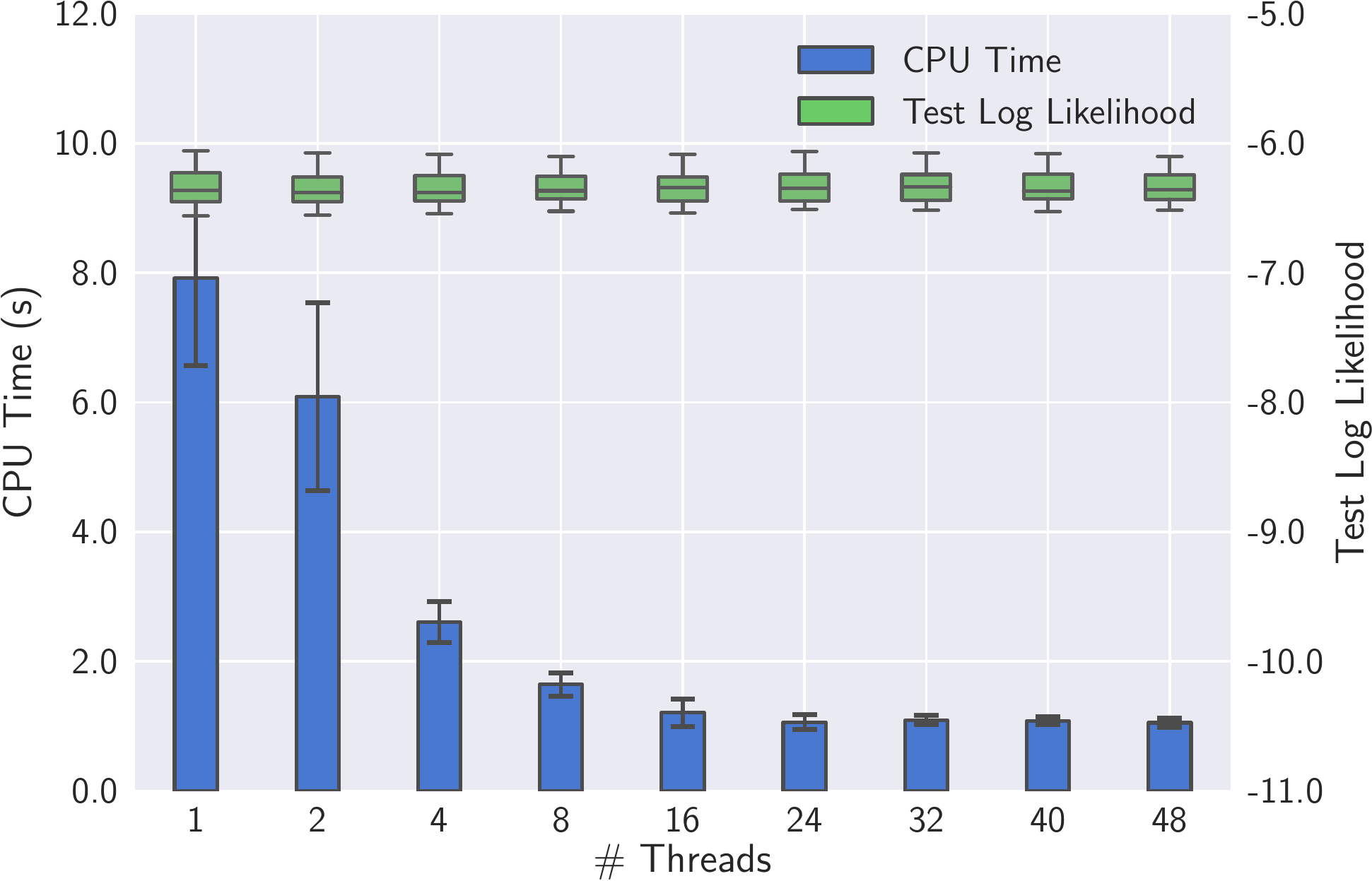}
    \captionsetup{font=scriptsize}
    \caption{SDA-DP with component ID}\label{fig:cputtll}
  \vspace{.4cm}
  \end{subfigure}
\hspace{.3cm}
\begin{subfigure}[b]{0.3\textwidth}
    \includegraphics[width=\textwidth]{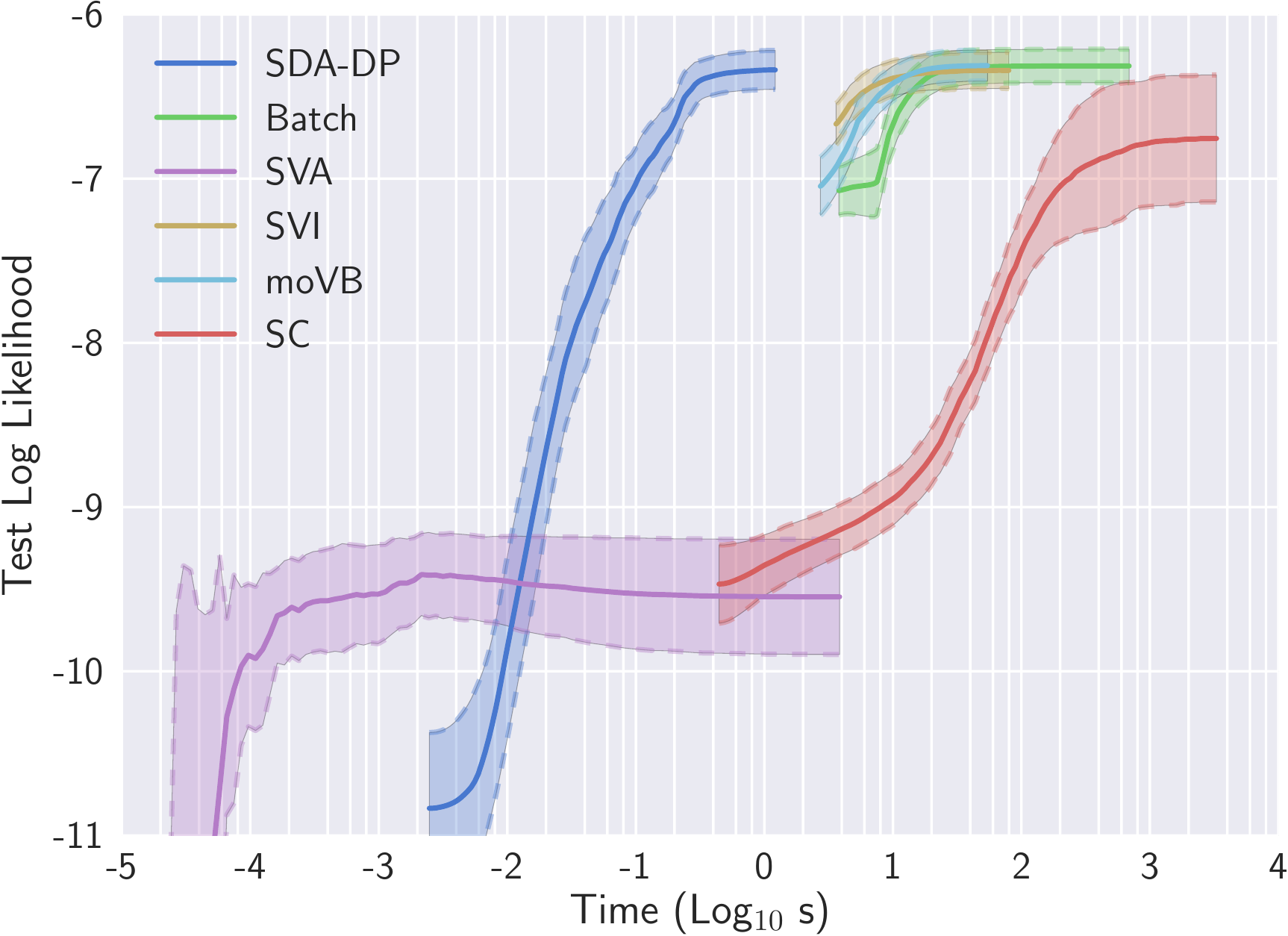} 
\captionsetup{font=scriptsize}
\caption{Test log-likelihood traces}\label{fig:testlltrace}
  \vspace{.4cm}
  \end{subfigure}
\begin{subfigure}[b]{0.32\textwidth}
    \includegraphics[width=\textwidth]{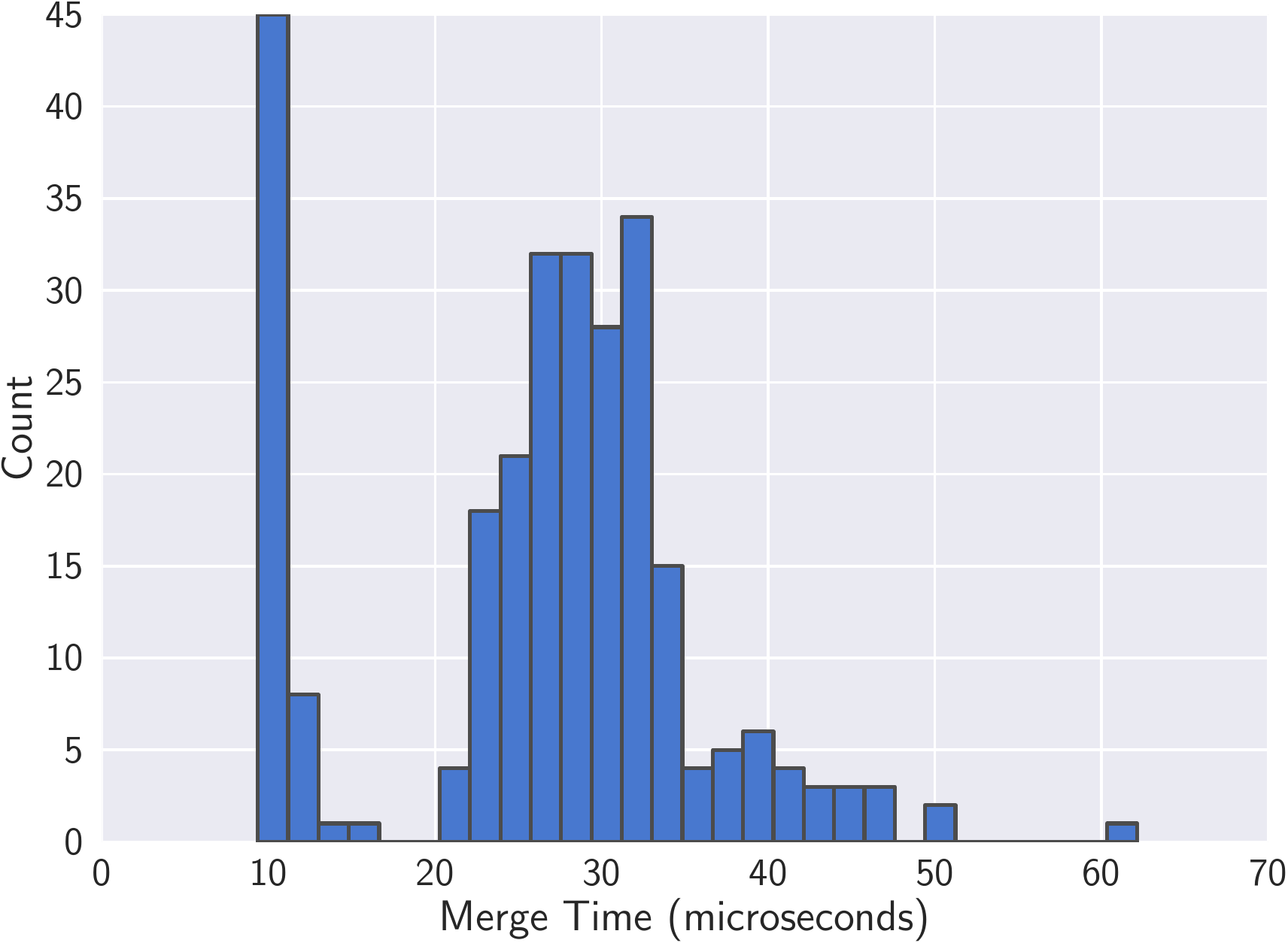} 
\captionsetup{font=scriptsize}
\caption{CPU time for component ID}\label{fig:mergehist}
  \end{subfigure}
\begin{subfigure}[b]{0.32\textwidth}
    \includegraphics[width=1.025\textwidth]{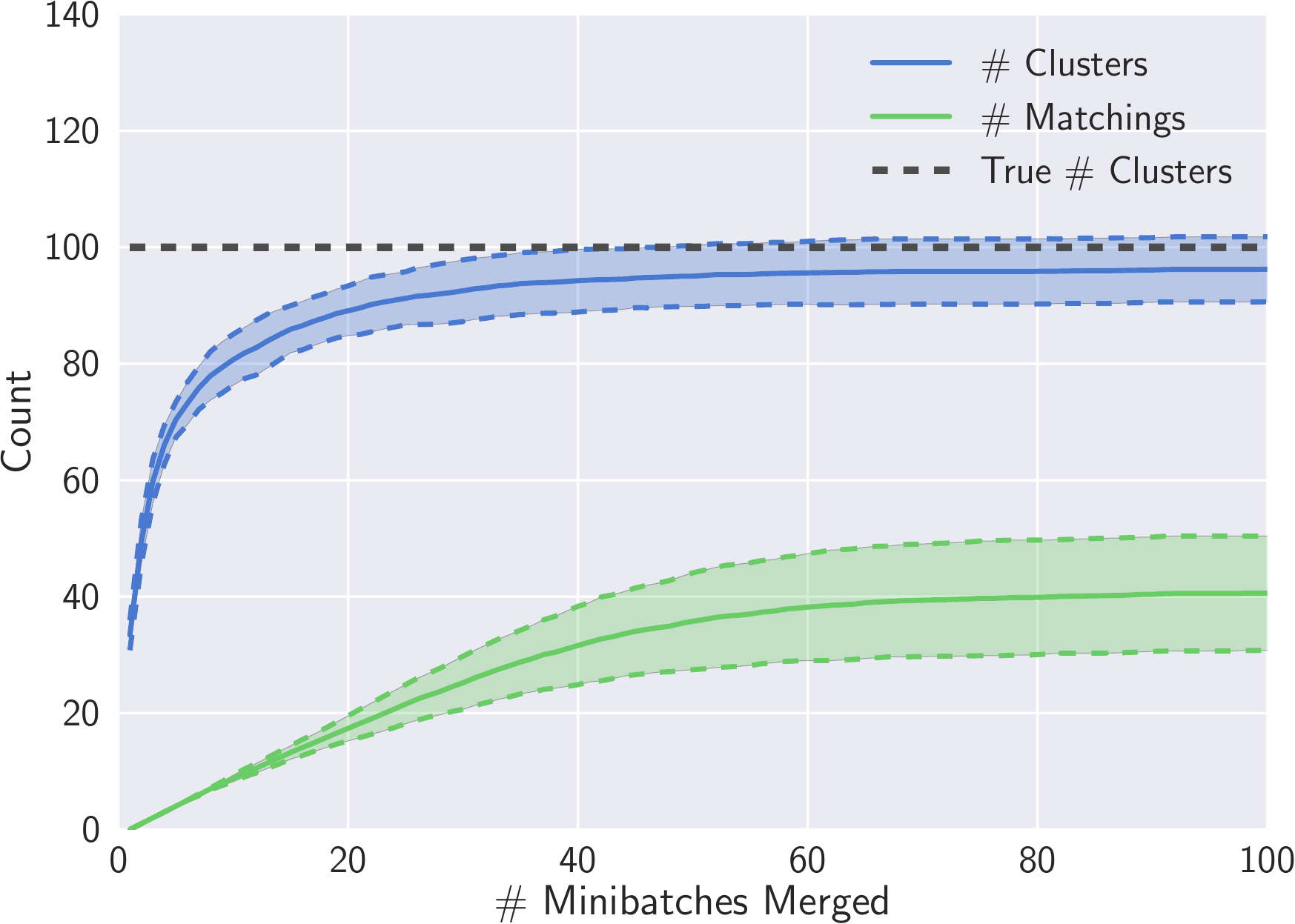} 
\captionsetup{font=scriptsize}
\caption{Cluster/component ID counts}\label{fig:clusmatchmini}
  \end{subfigure}
\begin{subfigure}[b]{0.315\textwidth}
    \includegraphics[width=1.03\textwidth]{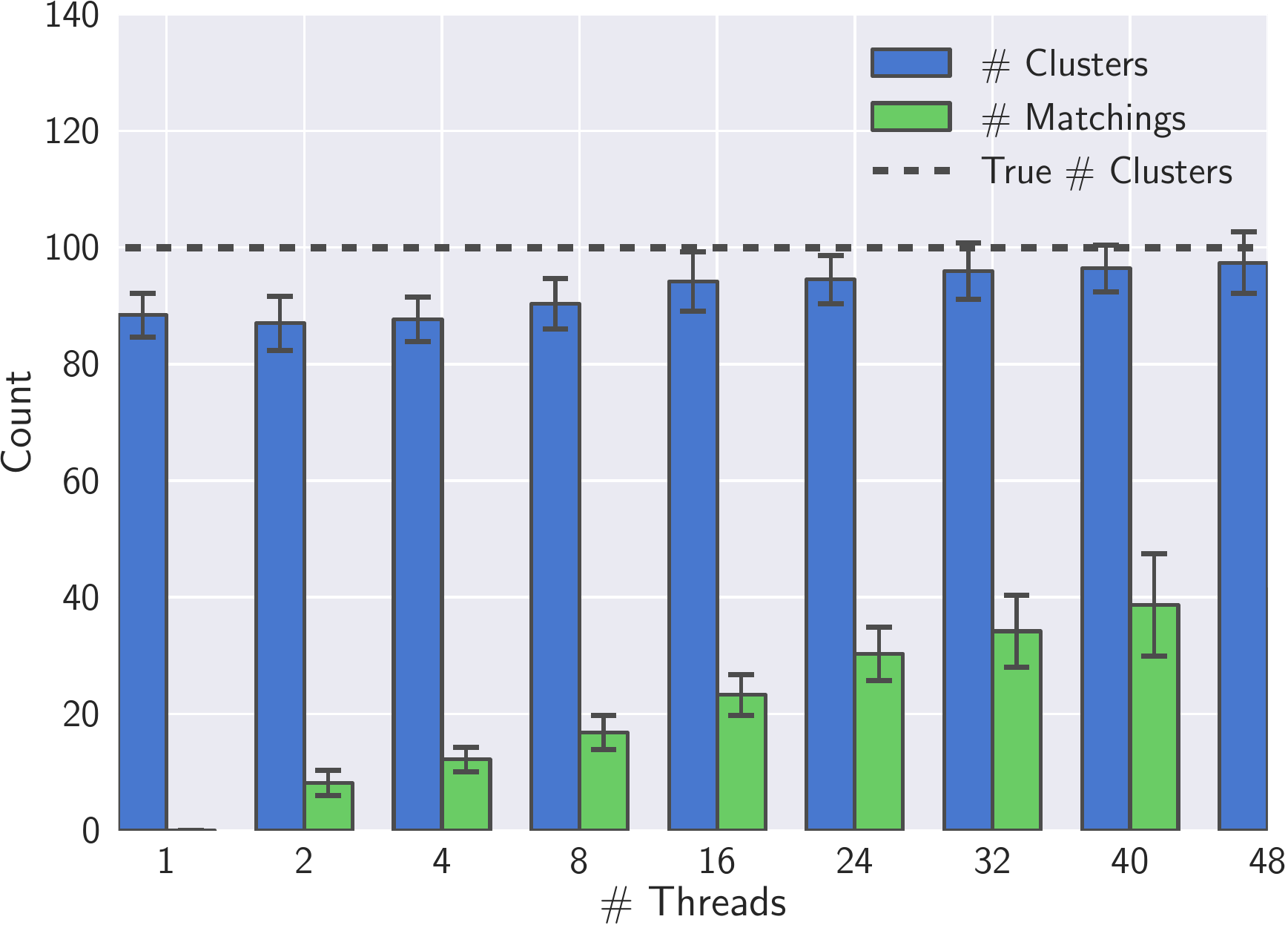}
    \captionsetup{font=scriptsize}
    \caption{Final cluster/component ID counts}\label{fig:clusmatchthr}
  \end{subfigure}
  \captionsetup{font=small}
  \caption{Synthetic results over 30 trials. (\ref{fig:synthbroken}-\ref{fig:cputtll}) Computation
    time and test log likelihood for SDA-DP with varying numbers of
    parallel threads, with component identification disabled
    (\ref{fig:synthbroken}) and enabled (\ref{fig:cputtll}).
    (\ref{fig:testlltrace}) Test log likelihood traces for SDA-DP (40 threads) and the comparison algorithms.
    (\ref{fig:mergehist}) Histogram of computation time
    (in microseconds) to solve the component identification
    optimization.  (\ref{fig:clusmatchmini}) Number of clusters and
    number of component identification problems solved as a function of
    the number of minibatch updates (40 threads).  (\ref{fig:clusmatchthr}) Final number of clusters and matchings
  solved with varying numbers of parallel threads.}\label{fig:s1}
\end{figure*}
In this section, the proposed inference framework is evaluated on the
DP Gaussian mixture with a normal-inverse-Wishart (NIW) prior. We
compare  the streaming, distributed procedure coupled with standard
variational inference~\citep{Blei06_BA} (SDA-DP) to five
state-of-the-art inference algorithms: memoized online variational
inference (moVB)~\citep{Hughes13_NIPS}, stochastic online variational
inference (SVI)~\citep{Hoffman13_JMLR} with learning rate
$(t+10)^{-\frac{1}{2}}$, sequential variational approximation
(SVA)~\citep{Lin13_NIPS} with cluster creation threshold $10^{-1}$ and
prune/merge threshold $10^{-3}$, subcluster splits MCMC (SC)~\citep{Chang13_NIPS}, and batch variational inference
(Batch)~\citep{Blei06_BA}. Priors were set by hand and all
methods were initialized randomly. Methods that use multiple passes
through the data (e.g.~moVB, SVI) were allowed to do so. moVB
was allowed to make birth/death moves, while SVI/Batch had fixed truncations.
All experiments were performed on a computer with 24 CPU cores and
12GiB of RAM. 

\textbf{Synthetic:} This dataset consisted 
of 100,000 2-dimensional vectors generated from a
Gaussian mixture model with 100 clusters and a $\mathrm{NIW}(\mu_0, \kappa_0,
\Psi_0, \nu_0)$ prior with $\mu_0 = 0$, $\kappa_0 = 10^{-3}$, $\Psi_0 =
I$, and $\nu_0 = 4$. The algorithms were given the true NIW prior, DP
concentration $\alpha = 5$, and minibatches of size 50. SDA-DP minibatch
inference was truncated to $K=50$ components, and all other algorithms
were truncated to $K=200$ components. 
Figure \ref{fig:s1}
shows the results from the experiment over 30 trials, which illustrate a number of important
properties of SDA-DP. First and foremost, ignoring the component identification
problem leads to decreasing model quality with increasing number of parallel
threads, since more matching mistakes are made (Figure \ref{fig:synthbroken}).
Second, if component identification is properly accounted for using the
proposed optimization, increasing the number of parallel threads
reduces execution time, but does not affect the final model quality
(Figure \ref{fig:cputtll}). Third, SDA-DP (with 40 threads) converges to 
the same final test log likelihood as the comparison algorithms in significantly
reduced time (Figure \ref{fig:testlltrace}). Fourth, each
component identification optimization typically takes $\sim10^{-5}$ seconds,
and thus matching accounts for less than a millisecond of total computation
and does not affect the overall computation 
time significantly (Figure \ref{fig:mergehist}). 
Fifth, the majority of the component matching problems are solved within the first 80
minibatch updates (out of a total of 2,000) -- afterwards, the true clusters have all been discovered
and the processing nodes contribute to those clusters rather than creating new
ones, as per the discussion at the end of Section \ref{sec:matchingopt} (Figure \ref{fig:clusmatchmini}). 
Finally, increased parallelization can be advantageous
in discovering the correct number of clusters; with only one thread, mistakes
made early on are built upon and persist, whereas with more threads there are
more component identification problems solved, and thus more chances to discover the
correct clusters (Figure \ref{fig:clusmatchthr}). 

\begin{figure*}[t!]
\centering
\begin{subfigure}[b]{0.28\textwidth}
  \raisebox{.3cm}{\includegraphics[width=\columnwidth]{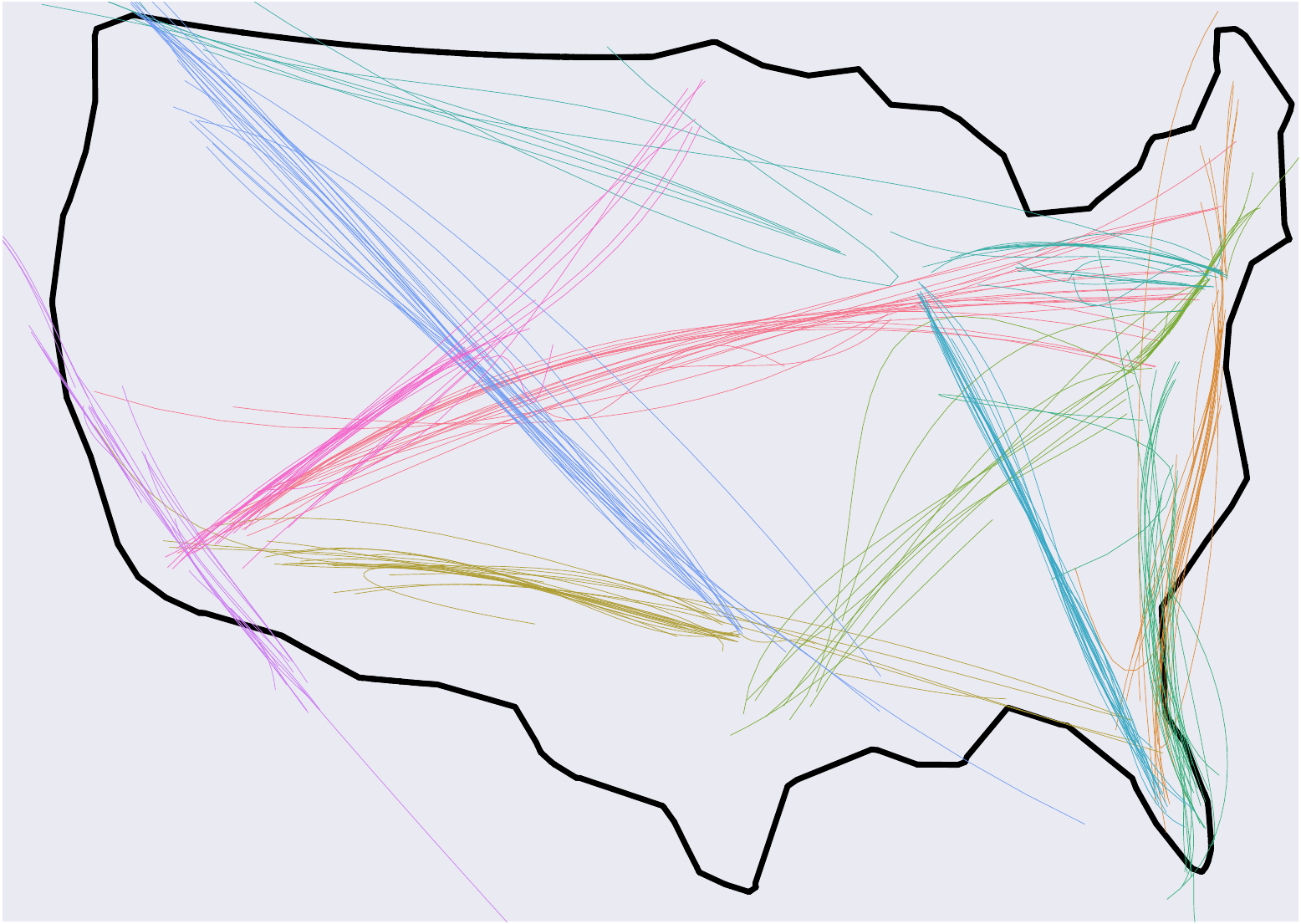}}
  \caption{Airplane trajectory clusters}\label{fig:planetraj}
\end{subfigure}
\begin{subfigure}[b]{0.31\textwidth}
\includegraphics[width=\columnwidth]{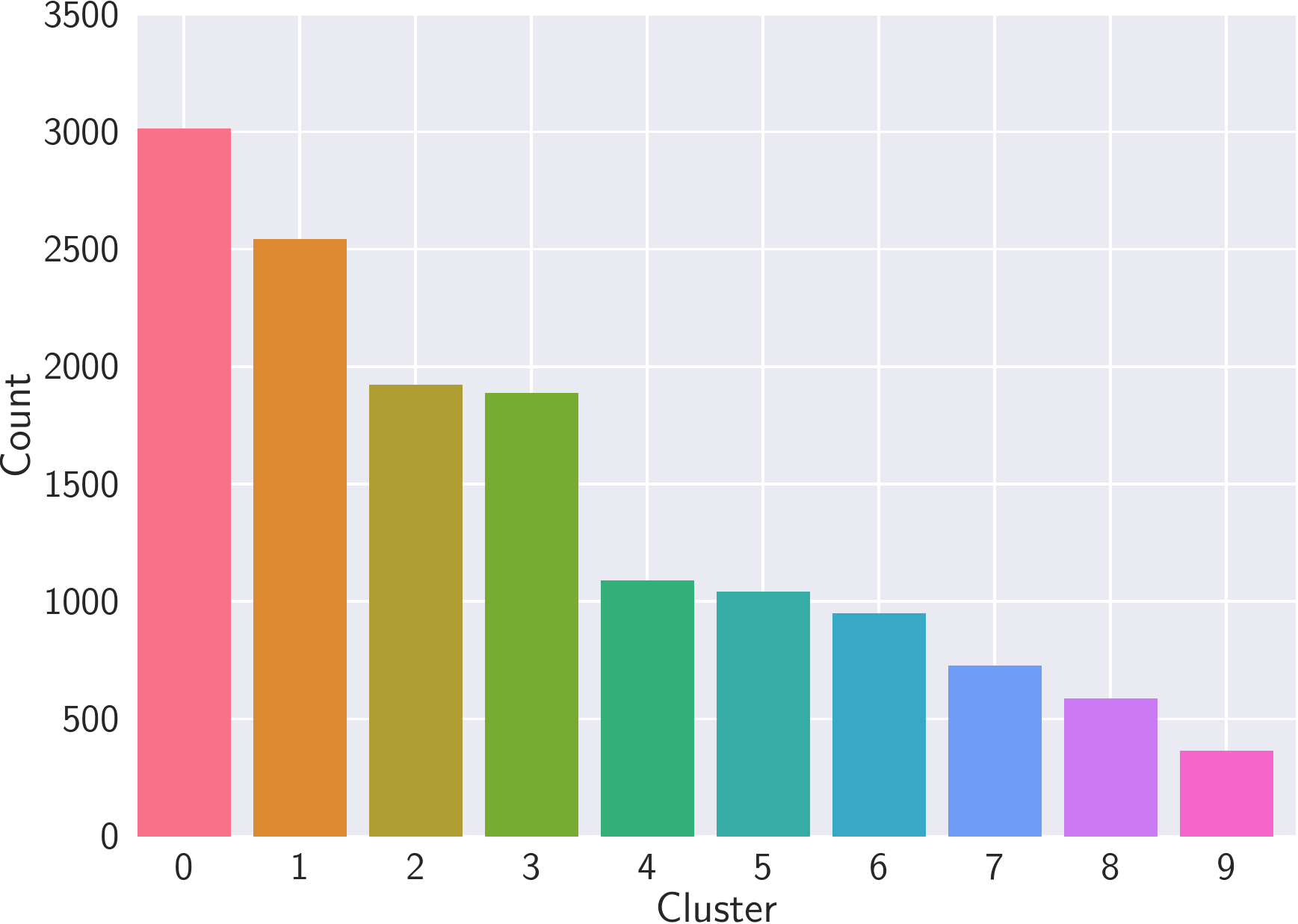}
  \caption{Airplane cluster weights}\label{fig:planect}
\end{subfigure}
\begin{subfigure}[b]{0.39\textwidth}
  \raisebox{.3cm}{\includegraphics[width=\textwidth]{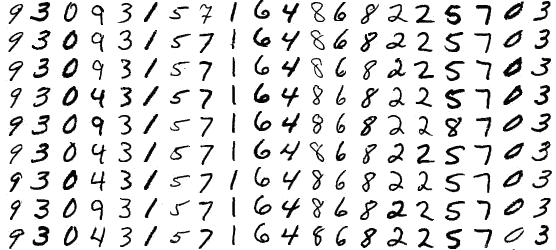}}
  \caption{MNIST clusters}\label{fig:mnist}
\end{subfigure}
\begin{subtable}[t]{\textwidth}
\vspace{.5cm}
  \centering
  \caption{Numerical results on Airplane, MNIST, and SUN}\label{fig:numres}
\vspace{-.2cm}
  {\small
\begin{tabular}{r|llllll}
& \multicolumn{2}{c}{\textbf{Airplane}} & \multicolumn{2}{c}{\textbf{MNIST}} &
\multicolumn{2}{c}{\textbf{SUN}}\\
	\textbf{Algorithm} & Time (s) & TestLL & Time (s) & TestLL & Time (s) & TestLL\\
	\hline
	SDA-DP & 0.66 & -0.55 & 3.0  &  -145.3  & 9.4 & -150.3 \\ 
	SVI & 1.50 & -0.59 & 117.4  & -147.1  & 568.9 & -149.9 \\
	SVA & 3.00 & -4.71 & 57.0 & -145.0  & 10.4 & -152.8 \\
	moVB & 0.69 & -0.72 & 645.9 & -149.2  & 1258.1 & -149.7\\
	SC & 393.6 & -1.06 & 1639.1 & -146.8 & 1618.4 & -150.6 \\
	Batch & 1.07 & -0.72 & 829.6 & -149.5 & 1881.5 & -149.7
	\end{tabular}
}
\end{subtable}

\captionsetup{font=small}
\caption{(\ref{fig:planetraj}-\ref{fig:planect}) Highest-probability instances
and counts for 10 trajectory clusters generated by SDA-DP. (\ref{fig:mnist}) Highest-probability instances for 20 clusters discovered by SDA-DP on MNIST. (\ref{fig:numres}) Numerical
results.}\label{fig:realresults}
\end{figure*}



\textbf{Airplane Trajectories:} This dataset consisted of $\sim$3,000,000 automatic
dependent surveillance broadcast (ADS-B) messages collected from 
planes across the United States during the period 2013-03-22
01:30:00UTC to 2013-03-28 12:00:00UTC. The messages were connected based on
plane call sign and time stamp, and erroneous trajectories
were filtered based on reasonable spatial/temporal bounds, yielding 15,022
trajectories with 1,000 held out for testing. The latitude/longitude points in
each trajectory were fit via linear regression, and the 3-dimensional parameter
vectors were clustered.
Data was split into minibatches of size 100, and SDA-DP used 16 parallel threads.

\textbf{MNIST Digits~\citep{mnist}:}
This dataset consisted of 70,000 $28\times28$ images of hand-written
digits, with 10,000 held out for testing. 
The images were reduced to 20 dimensions with PCA prior to clustering.
Data was split into minibatches of size 500, and SDA-DP used 48 parallel threads.

\textbf{SUN Images~\citep{sun397}:}
This dataset consisted of 108,755 images from 397 scene
categories, with 8,755 held out for testing. 
The images were 
reduced to 
20 dimensions with PCA prior to clustering. 
Data was split into minibatches of size 500, and SDA-DP used 48 parallel threads.

Figure \ref{fig:realresults} shows the results from the experiments on the three
real datasets. From a qualitative standpoint, SDA-DP discovers sensible clusters
in the data, as demonstrated in Figures \ref{fig:planetraj}--\ref{fig:mnist}. 
However, an important quantitative result is highlighted by Table \ref{fig:numres}: 
the larger a dataset is, the more the benefits of parallelism 
provided by SDA-DP become apparent. SDA-DP consistently provides a model
quality that is competitive with the other algorithms, but requires orders
of magnitude less computation time, corroborating similar findings
on the synthetic dataset.



%% file: conclusion.tex
\section{Conclusions}
This paper presented a streaming, distributed,
asynchronous inference algorithm for Bayesian nonparametric models,
with a focus on the combinatorial problem of matching minibatch
posterior components to central posterior components during asynchronous
updates. The main contributions are a component identification optimization
based on a minibatch posterior decomposition, a tractable bound on the 
objective for the Dirichlet process mixture, and experiments demonstrating
the performance of the methodology on large-scale datasets. While the present work focused on the DP mixture as a
guiding example, it is not limited to this model -- exploring 
the application of the proposed methodology to other BNP models is a potential
area for future research.

%

%

%% file: main.bbl
\begin{thebibliography}{26}
\providecommand{\natexlab}[1]{#1}
\providecommand{\url}[1]{\texttt{#1}}
\expandafter\ifx\csname urlstyle\endcsname\relax
  \providecommand{\doi}[1]{doi: #1}\else
  \providecommand{\doi}{doi: \begingroup \urlstyle{rm}\Url}\fi

\bibitem[Nobile(1994)]{Nobile94}
Agostino Nobile.
\newblock \emph{{{B}}ayesian Analysis of Finite Mixture Distributions}.
\newblock PhD thesis, Carnegie Mellon University, 1994.

\bibitem[Miller and Harrison(2013)]{Miller13_NIPS}
Jeffrey~W. Miller and Matthew~T. Harrison.
\newblock A simple example of {{D}}irichlet process mixture inconsistency for
  the number of components.
\newblock In \emph{Advances in Neural Information Processing Systems 26}, 2013.

\bibitem[Teh(2010)]{Teh10_EML}
Yee~Whye Teh.
\newblock {{D}}irichlet processes.
\newblock In \emph{Encyclopedia of Machine Learning}. Springer, New York, 2010.

\bibitem[Griffiths and Ghahramani(2005)]{Griffiths05_NIPS}
Thomas~L. Griffiths and Zoubin Ghahramani.
\newblock Infinite latent feature models and the {{I}}ndian buffet process.
\newblock In \emph{Advances in Neural Information Processing Systems 22}, 2005.

\bibitem[Broderick et~al.(2013)Broderick, Boyd, Wibisono, Wilson, and
  Jordan]{Broderick13_NIPS}
Tamara Broderick, Nicholas Boyd, Andre Wibisono, Ashia~C. Wilson, and
  Michael~I. Jordan.
\newblock Streaming variational {{B}}ayes.
\newblock In \emph{Advances in Neural Information Procesing Systems 26}, 2013.

\bibitem[Campbell and How(2014)]{Campbell14_UAI}
Trevor Campbell and Jonathan~P. How.
\newblock Approximate decentralized {{B}}ayesian inference.
\newblock In \emph{Proceedings of the 30th Conference on Uncertainty in
  Artificial Intelligence}, 2014.

\bibitem[Lin(2013)]{Lin13_NIPS}
Dahua Lin.
\newblock Online learning of nonparametric mixture models via sequential
  variational approximation.
\newblock In \emph{Advances in Neural Information Processing Systems 26}, 2013.

\bibitem[Zhang et~al.(2014)Zhang, Nott, Yau, and Jasra]{Nott14_JCGS}
Xiaole Zhang, David~J. Nott, Christopher Yau, and Ajay Jasra.
\newblock A sequential algorithm for fast fitting of {{D}}irichlet process
  mixture models.
\newblock \emph{Journal of Computational and Graphical Statistics}, 23\penalty0
  (4):\penalty0 1143--1162, 2014.

\bibitem[Hoffman et~al.(2013)Hoffman, Blei, Wang, and Paisley]{Hoffman13_JMLR}
Matt Hoffman, David Blei, Chong Wang, and John Paisley.
\newblock Stochastic variational inference.
\newblock \emph{Journal of Machine Learning Research}, 14:\penalty0 1303--1347,
  2013.

\bibitem[Wang et~al.(2011)Wang, Paisley, and Blei]{Wang11_AISTATS}
Chong Wang, John Paisley, and David~M. Blei.
\newblock Online variational inference for the hierarchical {{D}}irichlet
  process.
\newblock In \emph{Proceedings of the 11th International Conference on
  Artificial Intelligence and Statistics}, 2011.

\bibitem[Bryant and Sudderth(2009)]{Bryant09_NIPS}
Michael Bryant and Erik Sudderth.
\newblock Truly nonparametric online variational inference for hierarchical
  {{D}}irichlet processes.
\newblock In \emph{Advances in Neural Information Proecssing Systems 23}, 2009.

\bibitem[Wang and Blei(2012)]{Wang12_NIPS}
Chong Wang and David Blei.
\newblock Truncation-free stochastic variational inference for {{B}}ayesian
  nonparametric models.
\newblock In \emph{Advances in Neural Information Processing Systems 25}, 2012.

\bibitem[Hughes and Sudderth(2013)]{Hughes13_NIPS}
Michael Hughes and Erik Sudderth.
\newblock Memoized online variational inference for {{D}}irichlet process
  mixture models.
\newblock In \emph{Advances in Neural Information Processing Systems 26}, 2013.

\bibitem[Chang and {Fisher III}(2013)]{Chang13_NIPS}
Jason Chang and John {Fisher III}.
\newblock Parallel sampling of {{DP}} mixture models using sub-clusters splits.
\newblock In \emph{Advances in Neural Information Procesing Systems 26}, 2013.

\bibitem[Neiswanger et~al.(2014)Neiswanger, Wang, and Xing]{Neiswanger14_UAI}
Willie Neiswanger, Chong Wang, and Eric~P. Xing.
\newblock Asymptotically exact, embarassingly parallel {{MCMC}}.
\newblock In \emph{Proceedings of the 30th Conference on Uncertainty in
  Artificial Intelligence}, 2014.

\bibitem[Carvalho et~al.(2010)Carvalho, Lopes, Polson, and
  Taddy]{Carvalho10_BA}
Carlos~M. Carvalho, Hedibert~F. Lopes, Nicholas~G. Polson, and Matt~A. Taddy.
\newblock Particle learning for general mixtures.
\newblock \emph{{{B}}ayesian Analysis}, 5\penalty0 (4):\penalty0 709--740,
  2010.

\bibitem[Stephens(2000)]{Stephens00_JRSSB}
Matthew Stephens.
\newblock Dealing with label switching in mixture models.
\newblock \emph{Journal of the Royal Statistical Society: Series B},
  62\penalty0 (4):\penalty0 795--809, 2000.

\bibitem[Jasra et~al.(2005)Jasra, Holmes, and Stephens]{Jasra05_SS}
Ajay Jasra, Chris Holmes, and David Stephens.
\newblock {{M}}arkov chain {{M}}onte {{C}}arlo methods and the label switching
  problem in {{B}}ayesian mixture modeling.
\newblock \emph{Statistical Science}, 20\penalty0 (1):\penalty0 50--67, 2005.

\bibitem[Teh et~al.(2006)Teh, Jordan, Beal, and Blei]{Teh06_JASA}
Yee~Whye Teh, Michael~I. Jordan, Matthew~J. Beal, and David~M. Blei.
\newblock Hierarchical {{D}}irichlet processes.
\newblock \emph{Journal of the American Statistical Association}, 101\penalty0
  (476):\penalty0 1566--1581, 2006.

\bibitem[Doshi-Velez and Ghahramani(2009)]{DoshiVelez09_ICML}
Finale Doshi-Velez and Zoubin Ghahramani.
\newblock Accelerated sampling for the {{I}}ndian buffet process.
\newblock In \emph{Proceedings of the International Conference on Machine
  Learning}, 2009.

\bibitem[Dubey et~al.(2014)Dubey, Williamson, and Xing]{Dubey14_UAI}
Avinava Dubey, Sinead Williamson, and Eric Xing.
\newblock Parallel {{M}}arkov chain {{M}}onte {{C}}arlo for {{P}}itman-{{Y}}or
  mixture models.
\newblock In \emph{Proceedings of the 30th Conference on Uncertainty in
  Artificial Intelligence}, 2014.

\bibitem[Edmonds and Karp(1972)]{Edmonds72_ACM}
Jack Edmonds and Richard Karp.
\newblock Theoretical improvements in algorithmic efficiency for network flow
  problems.
\newblock \emph{Journal of the Association for Computing Machinery},
  19:\penalty0 248--264, 1972.

\bibitem[Pitman(1995)]{Pitman95_PTRF}
Jim Pitman.
\newblock Exchangeable and partially exchangeable random partitions.
\newblock \emph{Probability Theory and Related Fields}, 102\penalty0
  (2):\penalty0 145--158, 1995.

\bibitem[Blei and Jordan(2006)]{Blei06_BA}
David~M. Blei and Michael~I. Jordan.
\newblock Variational inference for {{D}}irichlet process mixtures.
\newblock \emph{{{B}}ayesian Analysis}, 1\penalty0 (1):\penalty0 121--144,
  2006.

\bibitem[Le{{C}}un et~al.()Le{{C}}un, Cortes, and Burges]{mnist}
Yann Le{{C}}un, Corinna Cortes, and Christopher~J.C. Burges.
\newblock {{MNIST}} database of handwritten digits.
\newblock Online: \url{yann.lecun.com/exdb/mnist}.

\bibitem[Xiao et~al.()Xiao, Hays, Ehinger, Oliva, and Torralba]{sun397}
Jianxiong Xiao, James Hays, Krista~A. Ehinger, Aude Oliva, and Antonio
  Torralba.
\newblock {{SUN}} 397 image database.
\newblock Online: \url{vision.cs.princeton.edu/projects/2010/SUN}.

\end{thebibliography}
